\documentclass{article}
\usepackage[preprint, nonatbib]{neurips_2023}
\usepackage[utf8]{inputenc} 
\usepackage[T1]{fontenc}    
\usepackage{hyperref}       
\usepackage{url}            
\usepackage{booktabs}       
\usepackage{amsfonts}       
\usepackage{nicefrac}       
\usepackage{microtype}      
\usepackage{xcolor}         

\usepackage{graphicx}
\usepackage[backend=biber, style=authoryear, natbib=true, maxcitenames=1]{biblatex}
\bibliography{references.bib}

\title{Mini-GPTs: Efficient Large Language Models through \\ Contextual Pruning}

\author{
  Tim Valicenti\thanks{https://github.com/tval2/contextual-pruning} \\
  Massachusetts Institute of Technology\\
  Cambridge, MA 02142 \\
  \texttt{tvalicen@mit.edu} \\
  \And
  Justice Vidal \\
  Massachusetts Institute of Technology\\
  Cambridge, MA 02142 \\
  \texttt{jmvidal@mit.edu} \\
  \AND
  Ritik Patnaik \\
  Massachusetts Institute of Technology\\
  Cambridge, MA 02142 \\
  \texttt{rik01@mit.edu} \\
}

\begin{document}

\maketitle

\begin{abstract}
In AI research, the optimization of Large Language Models (LLMs) remains a significant challenge, crucial for advancing the field's practical applications and sustainability. Building upon the foundational work of Professor Song Han's lab at MIT, this paper introduces a novel approach in developing Mini-GPTs via contextual pruning. Our methodology strategically prunes the computational architecture of traditional LLMs, like Phi-1.5, focusing on retaining core functionalities while drastically reducing model sizes. We employ the technique across diverse and complex datasets, including US law, Medical Q\&A, Skyrim dialogue, English-Taiwanese translation, and Economics articles. The results underscore the efficiency and effectiveness of contextual pruning, not merely as a theoretical concept but as a practical tool in developing domain-specific, resource-efficient LLMs. Contextual pruning is a promising method for building domain-specific LLMs, and this research is a building block towards future development with more hardware compute, refined fine-tuning, and quantization.
\end{abstract}

\section{Introduction \& Literature Review}
The advent of Large Language Models (LLMs) like GPT-4 has marked a paradigm shift in artificial intelligence, offering unparalleled capabilities in natural language processing. However, their extensive computational demands pose significant challenges, particularly in terms of cost, latency, emission concerns, and cloud dependence. This has spurred interest in model optimization techniques, notably model pruning, to address these challenges.

Model pruning, as explored by \cite{1510.00149} in “Deep Compression: Compressing Deep Neural Networks with Pruning, Trained Quantization and Huffman Coding”, has emerged as a promising avenue for reducing neural network sizes without substantially compromising their performance. This technique involves systematically removing non-critical weights from a network, thereby reducing its complexity, size, cost, and latency. Further advancements by \cite{1803.03635} in “The Lottery Ticket Hypothesis: Finding Sparse, Trainable Neural Networks” introduced the concept of identifying and training sparse subnetworks within larger models, suggesting that these 'lottery tickets' can achieve similar accuracy to their dense counterparts.

This paper examines the application of contextual pruning in creating Mini-GPTs, smaller yet efficient versions of existing LLMs. By analyzing and removing less critical weights specific to different domains, such as law, healthcare, and finance, we aim to maintain or enhance model performance while significantly reducing size and resource usage. This approach stacks with those designed by \cite{1510.00149} as synapse pruning (or connection pruning), quantization, and neural architecture search may done separately to our approach.

The initial motivation for pruning on context came from the realization that modern open-source LLMs are trained on broad datasets (e.g. Wikipedia, commercial-free books, and Reddit) but B2B users are only leveraging a small fraction of the information latent in the network that's relevant to their use case. By analogy, an LLM used at a hospital doesn't need to know options trading and Shakespeare - it just needs common sense, logical reasoning skills, and healthcare domain knowledge. 

\section{Methodology}
Our methodology for developing Mini-GPTs through contextual pruning primarily focused on linear layers, activation layers, and embedding layers. We also considered various datasets and models. This section highlights these choices. 

\subsection{Data}
\begin{table}[h]
\centering
\begin{tabular}{|c|c|c|}
\hline
Category & Size (text entries) & Source \\ \hline
General (used for testing only) & 4k & wikitext-2-raw-v1 \\ \hline
US Law & 10k & lexlms \\ \hline
Medical Q\&A & 15k & Laurent1/MedQuad-MedicalQnADataset \\ \hline
English-Taiwanese Translation & 311k & zetavg/coct-en-zh-tw-translations-twp-300k \\ \hline
Skyrim Full Transcript & 35k & sentiment-lexicon-skyrim \\ \hline
Economics Textbook & 6k & tinymlFP (economics\_text) \\ \hline
\end{tabular}
\caption{Overview of datasets used}
\label{data_table}
\end{table}

Our data collection focused on diverse domains to ensure a comprehensive evaluation of our pruning methodology - they are listed in Table \ref{data_table}. The belief is that the more dissimilar two datasets are, the more differences in neuron importance we'll find (and then therefor be able to prune).

\subsection{Initial Model Selection}
\begin{table}[h]
\centering
\begin{tabular}{|c|c|c|c|}
\hline
Model & HuggingFace & Size & Params \\ \hline
Phi-1.5 & microsoft/phi-1\_5 & 5437 MiB & 1.4B \\ \hline
Opt-1.3 & facebook/opt-1.3b & 5019 MiB & 1.3B \\ \hline
Llama-1.3 & princeton-nlp/Sheared-LLaMA-1.3B & 5144 MiB & 1.3B \\ \hline
\end{tabular}
\caption{Model selection}
\label{model_table}
\end{table}

We selected GPT-like architectures due to their robustness and popularity in various NLP tasks, including machine translation and multiple choice question answering. Our base models, highlighted in Table \ref{model_table}, are pre-trained transformers built by Microsoft (Phi-1.5) or Meta (Llama-1.3 and Opt-1.3), and they each came with a customized Byte-Pair Encoding (BPE) tokenizer in HuggingFace. 

\subsection{Contextual Analysis for Pruning}
 We conducted a detailed analysis of neuron outputs across linear layers, activation functions, and embeddings. This analysis helped us identify the weights that were less crucial for maintaining performance in specific domains. 
\\ \\
Contextual Analysis for Pruning: This crucial step involved three types of pruning, each targeting different model components:

\subsubsection{Linear Layer Pruning}
\begin{figure}[h]
\centering
\includegraphics[width=1\textwidth]{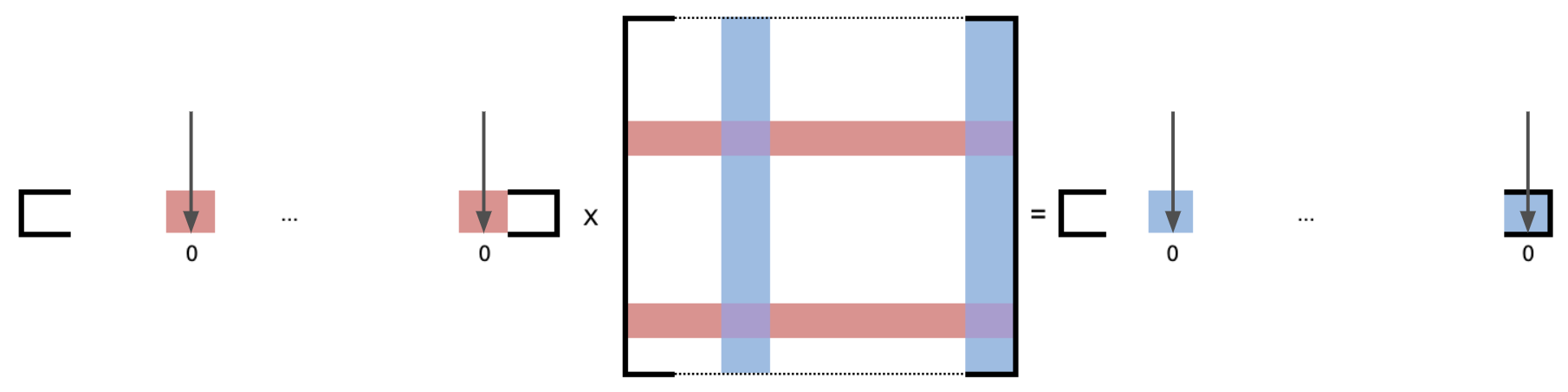}
\caption{Linear Layer Pruning}
\label{fig:linear_pruning}
\end{figure}

\begin{equation}
    m_j = \frac{1}{n} \sum_{b=1}^{n} || \mathbf{a}_{j,b} ||_1 < \epsilon_t
    \label{eq:pruning}
\end{equation}

To contextual prune the linear layers of an LLM, we tracked the neuron outputs and calculated, for each dataset, the normalized L1-norm of each neuron. Equation \ref{eq:pruning} shows this where $\mathbf{a}_{j,b}$ is the j-th neuron of batch $b$, $m
_j$ is the j-th activation's average magnitude across batches and $\epsilon_t$ is our pruning threshold. 

Figure \ref{fig:linear_pruning} conceptually shows how this impacts pruning by looking at a basic linear layer computation. When normalized across input batches, if the L1-norm is close to the pruning threshold then we prune the corresponding unused rows in the transpose weight matrix (red). Similarly, when normalized across output batches we identify which columns in the transpose weight matrix to prune (since they are not being utilities due to neuron-synapses interaction. 

\begin{figure}[h]
\centering
\includegraphics[width=0.8\textwidth]{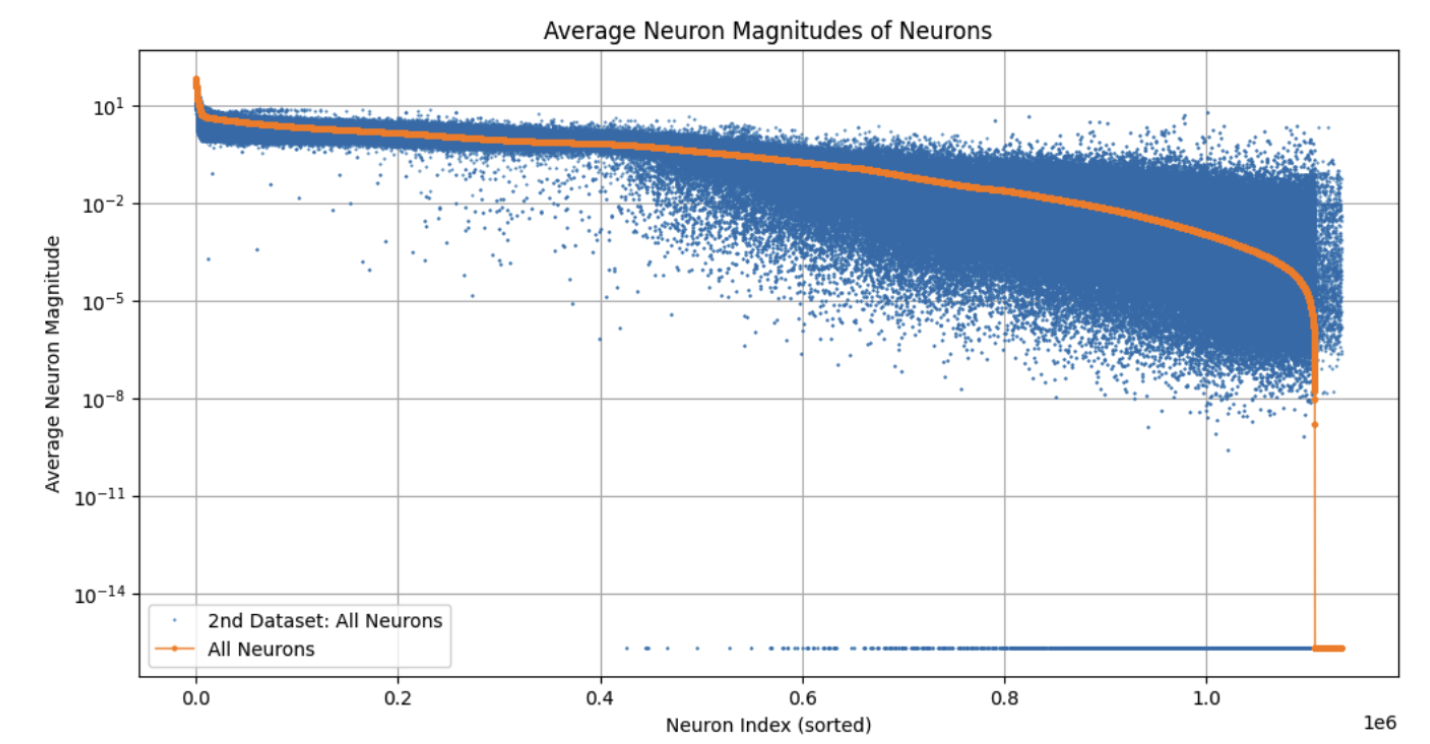}
\caption{comparison between magnitudes of neurons between skyrim and healthcare domains}
\label{fig:magnitude_plot}
\end{figure}

In Figure \ref{fig:magnitude_plot} we show example output of L1-norms for each neuron normalized for 2 datasets. Blue scatter points that fall below the orange line mean the neurons were activated more heavily in the first dataset as compared to the second dataset - and perhaps could be pruned from dataset 2.

\subsubsection{Activation Layer Pruning}
\begin{figure}[h]
\centering
\includegraphics[width=1.0\textwidth]{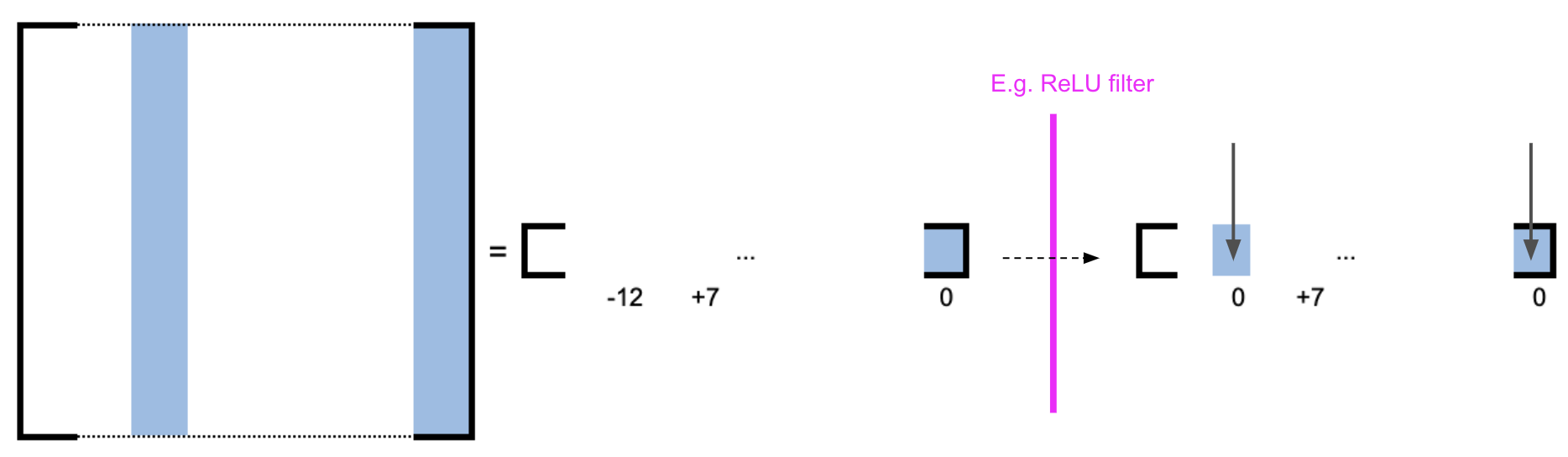}
\caption{Activation Layer Pruning}
\label{fig:activation_pruning}
\end{figure}
This pruning targeted the activation layers, where non-essential activation neurons are identified and removed. As shown in Figure \ref{fig:activation_pruning}, the approach is very similar to that of linear layers. One main difference is that we only look at the outputs of the layer, not the inputs. The other difference is that we must look to the previous layer to prune the weight from. If the normalized L1-norm of the activation neuron is below the pruning threshold then we prune the corresponding column in the transpose weight matrix of the prior layer. In the 3 models we looked at this was primarily done to GeLU and ReLU layers.

\subsubsection{Embedding Layer Pruning}
\begin{figure}[h]
\centering
\includegraphics[width=0.8\textwidth]{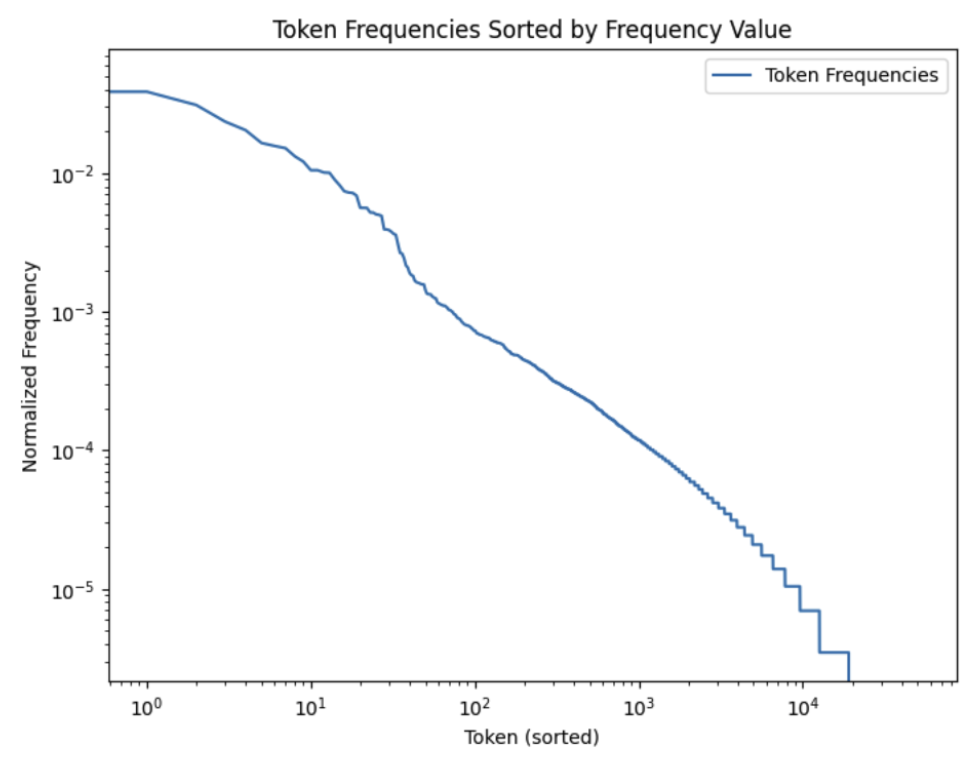}
\caption{Embedding Layer Pruning}
\label{fig:embedding_pruning}
\end{figure}

Lastly, we created functionality to prune embeddings layers (and the corresponding LM Head layer). This was done by measuring the token frequency of a particular dataset. While this approach works, we found that in order to use effectively very large calibration sets are needed to provide confidence that a token is truly not needed. One way to do this is to compare the token frequency curves of different domains.

\section{Evaluation and Results}
In this section, we present the evaluation methodology and results of our Mini-GPTs post contextual pruning. We used two primary metrics for evaluation: perplexity and multiple-choice question (MCQ) testing.
\subsection{Perplexity Evaluation}

\begin{table}[h]
\centering
\begin{tabular}{lrrrrr}
\toprule
Phi-1.5 &  Base &  Post prune &  Fine-tune & Recovery epochs & Relative Size (\%) \\
\midrule
Medical     &            4.640 &                  4.579 &                 2.722 &                1 &             90.134 \\
Skyrim      &           30.989 &                 29.728 &                12.687 &                1 &             89.805 \\
Economics   &           15.165 &                 15.132 &                 6.728 &                1 &             97.064 \\
Translation &           20.292 &                 20.198 &                10.429 &                1 &             97.765 \\
Legal       &           20.029 &                 19.904 &                 8.490 &                1 &             94.490 \\
\bottomrule
Opt-1.3 & {} & {} & {} & {} & {} \\
\midrule
Medical     &            3.829 &                  4.615 &                 3.203 &                1 &             88.369 \\
Skyrim      &           19.777 &                 26.836 &                 8.373 &                1 &             89.820 \\
Economics   &           13.283 &                 16.916 &                 8.639 &                1 &             91.225 \\
Translation &           17.187 &                 26.630 &                11.994 &                2 &             90.619 \\
Legal       &           14.251 &                 17.260 &                11.444 &                1 &             90.427 \\
\bottomrule
Llama-1.3 & {} & {} & {} & {} \\
\midrule
Medical     &            3.177 &                  3.177 &                 1.799 &                1 &             99.789 \\
Skyrim      &           15.712 &                 15.705 &                 4.612 &                1 &             99.717 \\
Economics   &            8.514 &                  8.513 &                 3.535 &                1 &             99.760 \\
Translation &           14.607 &                 14.606 &                 5.065 &                1 &             99.841 \\
Legal       &            8.312 &                  8.312 &                 3.613 &                1 &             99.765 \\
\bottomrule
\end{tabular}
\caption{Perplexity results of pruning models with linear and activation threshold of $10^{-3}$ and pruning embeddings <= 0; Models fine-tuned until perplexity recovered, with max training epochs of 200.}
\label{perplexity_table}
\end{table}

Perplexity measures how well a large language model can predict the next word given a string of context and is a standard metric in determining a language model's performance. Generally, a lower perplexity indicates a better model. From Table \ref{perplexity_table}, we generally observe a reduction or no change in perplexity across all datasets post-pruning and fine-tuning, indicating that the models were able to retain much of their ability in their respective domains despite the reduction in usable parameters.

\subsection{Multiple-Choice Question Testing}
We further evaluated our models on 100 domain-specific MCQs to further ensure that the model retained its ability prior to pruning. Since only phi-1.5 could generate a string containing the correct answer choice, to remain consistent across each model, a model's answer to a MCQ was selected by picking the question + answer string concatenation that resulted in the lowest perplexity, effectively using the model's best guess. The results shown in Table \ref{question_table} that the pruned models performed comparably and, in some cases, better than their un-pruned versions, demonstrating the effectiveness of our pruning methodology.

\begin{table}[h]
\centering
\begin{tabular}{lrrrrr}
\toprule
Phi-1.5 & Base (\%) &  Post prune (\%) &  Fine-tune (\%) & Recovery epochs & Relative Size (\%) \\
\midrule
Medical     &         33.000 &               27.000 &              25.000 &                1 &             90.134 \\
Skyrim      &         62.000 &               63.000 &              63.000 &                1 &             89.805 \\
Economics   &         68.421 &               67.368 &              68.421 &                1 &             97.064 \\
Translation &         36.000 &               37.000 &              38.000 &                1 &             97.765 \\
\bottomrule
Opt-1.3 & {} & {} & {} & {} \\
\midrule
Medical     &         32.000 &               25.000 &              24.000 &                1 &             88.369 \\
Skyrim      &         73.000 &               58.000 &              67.000 &                1 &             89.820 \\
Economics   &         46.316 &               47.368 &              51.579 &                1 &             91.225 \\
Translation &         38.000 &               35.000 &              32.000 &                2 &             90.619 \\
\bottomrule
Llama-1.3 & {} & {} & {} & {} \\
\midrule
Medical     &         30.000 &               30.000 &              31.000 &                1 &             99.789 \\
Skyrim      &         65.000 &               65.000 &              63.000 &                1 &             99.717 \\
Economics   &         48.421 &               49.474 &              46.316 &                1 &             99.760 \\
Translation &         46.000 &               46.000 &              53.000 &                1 &             99.841 \\
\bottomrule
\end{tabular}
\caption{MCQ accuracy results of pruning models with linear and activation threshold of $10^{-3}$ and pruning embeddings <= 0; Models fine-tuned until perplexity recovered, with max training epochs of 200.}
\label{question_table}
\end{table}

\subsection{Large Pruning Threshold}
To test the limits of our pruning methodology, we also tested a linear and activation threshold of $10^{-1}$.

\begin{table}[h]
\centering
\begin{tabular}{lrrrrr}
\toprule
Phi-1.5 &  Base &  Post prune &  Fine-tune &  Recovery epochs &  Relative Size (\%) \\
\midrule
Medical     &            4.640 &              35417.938 &                 4.312 &               25 &             58.116 \\
Skyrim      &           30.989 &              20174.240 &                27.963 &               21 &             59.808 \\
Economics   &           15.165 &              25619.248 &                11.178 &               13 &             66.972 \\
Translation &           20.292 &                129.540 &                13.671 &                5 &             69.069 \\
Legal       &           20.029 &              18902.793 &                18.519 &               11 &             64.410 \\
\bottomrule
Opt-1.3 & {} & {} & {} & {} & {} \\
\midrule
Medical     &            3.829 &               9559.019 &                22.407 &              200 &             64.703 \\
Skyrim      &           19.777 &               1830.905 &                19.774 &               71 &             64.412 \\
Economics   &           13.283 &               7515.678 &                37.525 &              200 &             64.957 \\
Translation &           17.187 &               5248.911 &                36.943 &              200 &             63.334 \\
Legal       &           14.251 &               7545.842 &                45.976 &              200 &             65.091 \\
\bottomrule
Llama-1.3 &  {} &  {} &  {} & {} & {} \\
\midrule
Medical     &            3.177 &              69290.547 &                 3.342 &              200 &             69.126 \\
Skyrim      &           15.712 &               3364.670 &                13.635 &               33 &             68.098 \\
Economics   &            8.514 &              71864.391 &                 8.403 &               85 &             68.868 \\
Translation &           14.607 &              53817.781 &                14.074 &               78 &             69.451 \\
Legal       &            8.312 &              16954.877 &                 8.204 &               45 &             69.513 \\
\bottomrule
\end{tabular}
\caption{Perplexity results of pruning models with linear and activation threshold of $10^{-1}$ and pruning embeddings <= 0; Models fine-tuned until perplexity recovered, with max training epochs of 200}
\label{large_perplexity_table}
\end{table}
From Table \ref{large_perplexity_table}, we find a potential size reduction of up to 41.884\% with the Phi model while recovering perplexity prior to pruning. Generally, however, the results indicate we are approaching the limit of pruning for these models as Opt struggles heavily to recover perplexity prior to pruning, and Phi and Llama take 10s of epochs to recover where only 1 was necessary in the $10^{-3}$ case.
Furthermore, looking at the MCQ results[\ref{large_question_table}] for each model, overall, we find that accuracy decreases again after fine-tuning while the perplexity on the fine-tuning set decreases, indicating overfitting. Further testing is required to determine if this can be mitigated with a larger, more representative dataset for each category or if this level of size reduction is too great entirely.
The results on the much larger English to Taiwanese dataset suggest the former, as MCQ accuracy increased across all models after fine-tuning.
\begin{table}[h]
\centering
\begin{tabular}{lrrrrr}
\toprule
Phi-1.5 & Base (\%) & Post prune (\%) & Fine-tune (\%) &  Recovery epochs &  Relative Size (\%) \\
\midrule
Medical     &         33.000 &               25.000 &              25.000 &               25 &             58.116 \\
Skyrim      &         62.000 &               28.000 &              32.000 &               21 &             59.808 \\
Economics   &         68.421 &               35.789 &              29.474 &               13 &             66.972 \\
Translation &         36.000 &               30.000 &              33.000 &                5 &             69.069 \\
\bottomrule
Opt-1.3 & {} & {} & {} & {} & {} \\
\midrule
Medical     &         32.000 &               32.000 &              28.000 &              200 &             64.703 \\
Skyrim      &         73.000 &               27.000 &              23.000 &               71 &             64.412 \\
Economics   &         46.316 &               29.474 &              21.053 &              200 &             64.957 \\
Translation &         38.000 &               30.000 &              31.000 &              200 &             63.334 \\
\bottomrule
Llama-1.3 &  {} &  {} &  {} & {} & {} \\
\midrule
Medical     &         30.000 &               25.000 &              24.000 &              200 &             69.126 \\
Skyrim      &         65.000 &               27.000 &              30.000 &               33 &             68.098 \\
Economics   &         48.421 &               21.053 &              17.895 &               85 &             68.868 \\
Translation &         46.000 &               26.000 &              28.000 &               78 &             69.451 \\
\bottomrule
\end{tabular}
\caption{MCQ accuracy results of pruning models with linear and activation threshold of $10^{-1}$ and pruning embeddings <= 0; Models fine-tuned until perplexity recovered, with max training epochs of 200}
\label{large_question_table}
\end{table}

\section{Conclusion and Future Work}
Our research on Mini-GPTs through contextual pruning has shown promising results in balancing efficiency with performance. The significant reduction in model sizes, coupled with maintained or improved accuracy in domain-specific tasks, validates our approach. For future work, we plan to focus on several key areas:

\begin{itemize}
  \item \textbf{Pruning off Max Neuron Magnitude}: We aim to explore pruning based on maximum neuron magnitude, which might be more robust against outliers.
  \item \textbf{Fine Tune and Evaluate on Larger Datasets}: To enhance representativeness and generalizability, we will fine tune our models on larger datasets and more compute power to prevent overfitting.
  \item \textbf{Combining with Other Optimization Techniques}: We plan to integrate our pruning method with techniques like quantization for higher performing models.
  \item \textbf{Exploring Other Models}: Our methodology will be applied to more up-to-date models, such as Phi-2 by Microsoft.
\end{itemize}

Our research opens new avenues in domain-specific model optimization, promising wider applications for LLMs in the world. This especially allows for more on-prem usage in industries such as gaming, healthcare, defense, and consumer use.

\medskip
\small
\printbibliography


\end{document}